\documentclass[manuscript]{acmart}
\usepackage[T1]{fontenc}
\usepackage{multirow}
\begin{document}
\title{Addressing bias in Recommender Systems: A Case Study on Data Debiasing Techniques in Mobile Games}

\author{Yixiong Wang}
\affiliation{%
  \institution{King}
  \city{Stockholm}
  \country{Sweden}}
\email{wyx.ei.99@gmail.com}

\author{Maria Paskevich}
\authornote{Corresponding author.}
\affiliation{%
  \institution{King}
  \city{Stockholm}
  \country{Sweden}}
\email{maria.paskevich@king.com}

\author{Hui Wang}
\affiliation{%
  \institution{King}
  \city{Stockholm}
  \country{Sweden}}
\email{madeleine.renstrom@king.com}

\renewcommand{\shortauthors}{Wang et al.}

% \author{Yixiong Wang\inst{1},
% Maria Paskevich\inst{2} 
% and Hui Wang\inst{3}}
%
%\authorrunning{Wang, Paskevich and Wang}
%#\institute{King, Stockholm, Sweden 
%\email{wyx.ei.99@gmail.com}
%\and
%King, Stockholm, Sweden
%\email{maria.paskevich@king.com}, corresponding author
%\and
%King, Stockholm, Sweden
%\email{madeleine.renstrom@king.com}}

\begin{abstract}
The mobile gaming industry, particularly the free-to-play sector, has been around for more than a decade, yet it still experiences rapid growth. The concept of games-as-service requires game developers to pay much more attention to recommendations of content in their games. With recommender systems (RS), the inevitable problem of bias in the data comes hand in hand. A lot of research has been done on the case of bias in RS for online retail or services, but much less is available for the specific case of the game industry. Also, in previous works, various debiasing techniques were tested on explicit feedback datasets, while it is much more common in mobile gaming data to only have implicit feedback. This case study aims to identify and categorize potential bias within datasets specific to model-based recommendations in mobile games, review debiasing techniques in the existing literature, and assess their effectiveness on real-world data gathered through implicit feedback. The effectiveness of these methods is then evaluated based on their debiasing quality, data requirements, and computational demands. 

\end{abstract}


\maketitle              % typeset the header of the contribution
\section{Introduction}
In the context of mobile gaming, delivery of content to players through recommendations plays an important role. It could include elements such as, for example, in-game store products or certain parts of content. However, RSs used within this context are susceptible to bias due to (1) limited exposure: unlike in webshops (e.g. Amazon), available placements for sellable products in mobile games are often limited, and showing one product to a user means that alternatives would not be displayed; (2) the common approach of segmenting content through fixed heuristics before adopting RS introduces biases in the training data, which influences the development of these models. Traditionally, at King we have been addressing these biases by either training models on biased data, or by establishing holdout groups of users who would receive random recommendations for a period of time in order to collect a uniform dataset that reflects user preference in an unbiased way. Although the second approach allows the collection of unbiased data, it could compromise user experience for a segment of players, and may lead to significant operational costs and potential revenue losses. In previous studies, researchers have primarily focused on data derived from explicit feedback, where users rate items using a numerical scale, and various debiasing techniques are tested on this data. However, within the realm of mobile gaming, obtaining explicit feedback affects from user experience, making it challenging to collect. As an alternative, data is often collected through implicit feedback \cite{oardimplicit}, where user preferences are inferred from behaviors such as impressions, purchases, and other interactions. Given these challenges, our objectives in this study are: (1) to identify and categorize potential bias within our datasets; (2) to conduct a review of existing literature on debiasing techniques and assess their effectiveness on publicly available datasets;  (3) to adapt and apply debiasing strategies, originally developed for explicit feedback data, to the implicit feedback data specific to King, and (4) to evaluate and compare the efficacy of different methods based on the quality of debiasing, data requirements, and computational complexity.
\section{Related work}
The existing literature on addressing debiasing techniques in RS presents a well-structured and categorized list of methodologies \cite{chen2021bias}\cite{steck2010mnar}. It suggests that the selection of particular debiasing techniques should depend on the specific types of bias present in the data, as well as on the availability of unbiased data samples. In recommender systems for mobile games, various types of bias can arise, including but not limited to selection bias, exposure bias, position bias, and conformity bias. Some of the relevant methods to debias the data in these cases could be The Inverse Propensity Scoring (IPS) \cite{IPS2021} method, which deals with selection and exposure biases by weighting observations inversely to their selection probability, and does so without need for unbiased data. Yet the method could potentially result in high variance due to the challenges in accurately estimating propensities. Potential solutions to the high variance issue of IPS method include, for example, using Doubly Robust (DR) learning \cite{DRPostClick2021} that introduces a novel approach to loss functions as a combination of IPS-based models with imputation-based models. The combination of two models assures doubly robustness property when either of the two components (propensity estimation or imputed data) remains accurate. This method, though, relies on having an unbiased data sample to work. Another option is model-agnostic and bias-agnostic solutions like AutoDebias \cite{AutoDebias}, which are based on meta-learning to dynamically assign weights within the RS, aiming to neutralize biases across the board. A potential benefit of such solution is that it doesn't require knowing the types of bias present in the data, but as a downside, it also relies on randomized samples. In addition, the process of fitting multiple models makes training more computationally demanding.
Despite the advances and variety of available debiasing techniques, applying Recommendation Systems to mobile gaming content remains a relatively untapped area, with most of the publications focusing on building recommendations \cite{ContextualItemAwareRec} \cite{PersonalBundleRec} \cite{RecAppsEA}, and not on issues of imbalance and bias. Previous efforts at King introduced DFSNet \cite{DFSNet}, an end-to-end model-specific debiasing technique that enables training an in-game recommender on an imbalanced dataset without randomized data. 
This work aims to enrich King's debiasing toolkit by exploring model-agnostic solutions, specifically focusing on the challenges of content recommendations within mobile games. However, the architecture of DFSNet is complex, involving multiple modules, which can make the implementation and maintenance challenging. Moreover, it requires constant feedback loops over time and the model's performance is highly dependent on the quality and recency of the training data. %Furthermore, implicit feedback loop need to investigate further from this previous study. %

\begin{table}[!htpb]
    \centering
    \begin{minipage}{.4\textwidth}
        \centering
        \begin{figure}[H]  % Use the float package if H gives errors
            \includegraphics[width=\linewidth]{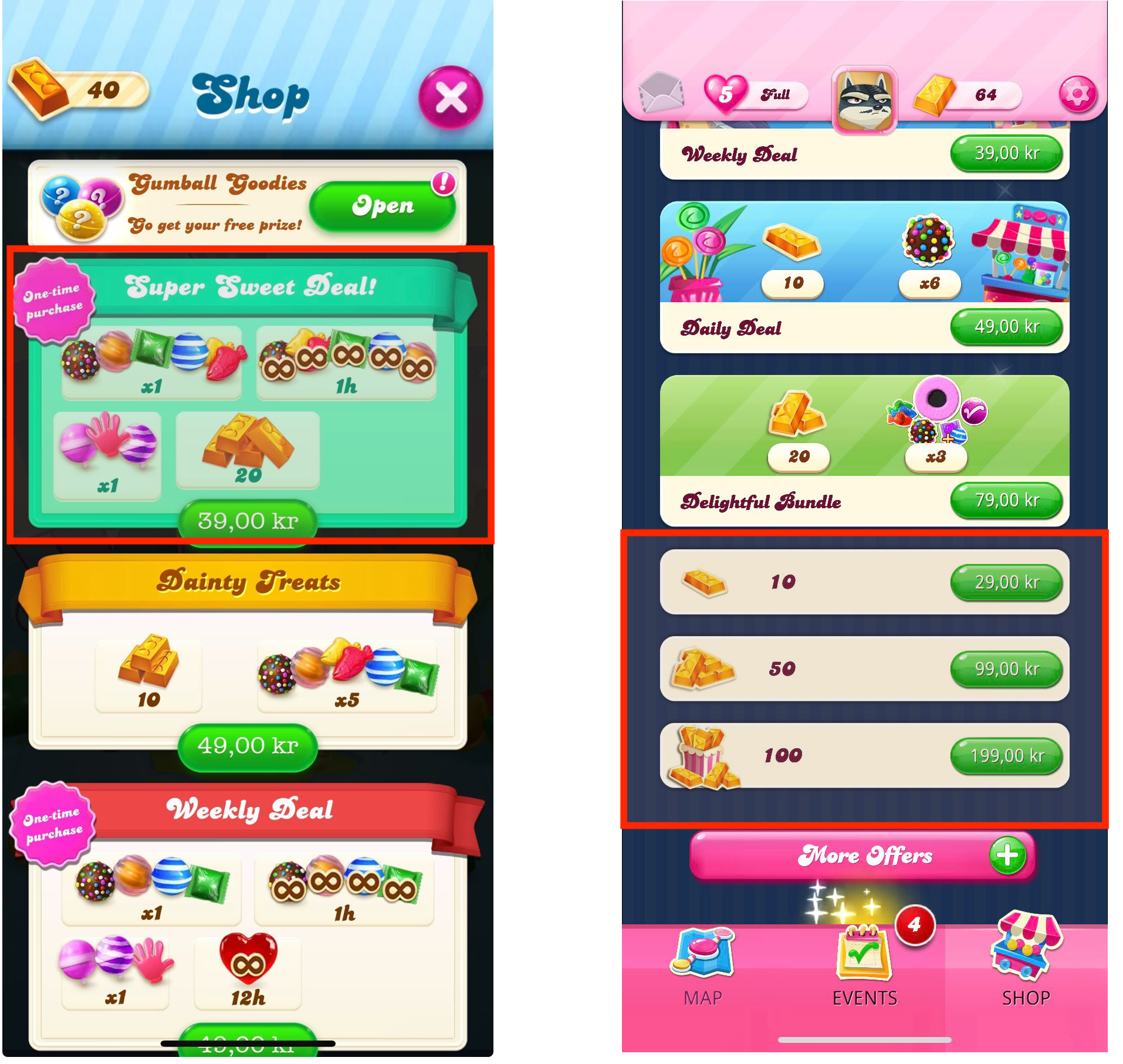}
            \caption{\small{Examples of content placements in Candy Crush Soda Saga (left) and Candy Crush Saga (right), highlighting biases: selection bias with a prominently placed product (left) and exposure bias with limited visibility, where products are hidden behind the "More Offers" button (right).}}
            \label{fig:soda}
        \end{figure}
    \end{minipage}\hfill
    \begin{minipage}{.55\textwidth}
    \caption{The sizes and feedback types of all datasets used in this study.A key difference is that the open datasets (COAT and YahooR3!) provide explicit feedback, while the proprietary datasets (A, B, and C) offer only implicit feedback (purchase/no purchase). Set A, a proprietary dataset, lacks randomized data, limiting debiasing options.}
        \centering
        \begin{tabular}{l c c c}
        \toprule
        Dataset & Biased samples & Unbiased samples & Feedback type \\
        \midrule
        COAT & 311k & 54k & Explicit\\
        yahooR3! & 12.5k & 75k & Explicit\\
        Set A & 47.6k & - & Implicit\\
        Set B & 100k & 218k& Implicit\\
        Set C & 980k & 1.2mln& Implicit\\
        \bottomrule
        \end{tabular}
        \label{tab:sample_sizes}
    \end{minipage}
\end{table}

% \begin{figure}[!htpb]
%     \centering
%     \begin{minipage}{.4\textwidth}
%         \centering
%         \includegraphics[width=\linewidth]{css_ccss.png}
%         \caption{\small{Examples of placements for recommended content in Candy Crush Soda Saga (left) and Candy Crush Saga (right).}}
%         \label{fig:soda}
%     \end{minipage}\hfill
%     \begin{minipage}{.55\textwidth}
%         \centering
%         \begin{tabular}{l c c c}
%         \toprule
%         Dataset & Biased samples & Unbiased samples & Feedback type \\
%         \midrule
%         COAT & 311k & 54k & Explicit\\
%         yahooR3! & 12.5k & 75k & Explicit\\
%         Set A & 47.6k & - & Implicit\\
%         Set B & 100k & 218k& Implicit\\
%         Set C & 980k & 1.2mln& Implicit\\
%         \bottomrule
%         \end{tabular}
%         \caption{Sizes of all datasets and their feedback types used in the study.}
%         \label{tab:sample_sizes}
%     \end{minipage}
% \end{figure}

\section{Methodology}
\subsection{Datasets}
Our study utilized two public datasets (COAT\cite{IPS2021}, yahooR3!\cite{yahoor3}) to validate theoretical results and three proprietary datasets from King (Set A, Set B, Set C) that are focused on user-item interactions in game shops within Match-3 Game A and Match-3 Game B (Fig.\ref{fig:soda}). The sizes of each dataset, along with their respective feedback types, are provided in Table  \ref{tab:sample_sizes}. We aimed to observe the effectiveness of different techniques on datasets collected with explicit feedback (public datasets), and those with implicit feedback (King's datasets). Explicit feedback is typically collected by asking users to rate items on a numerical scale, for example from 1 to 5, where 1 indicates disinterest, 2 signifies dissatisfaction, and 5 shows a preference. In contrast, Implicit feedback (as in the proprietary datasets) involves a binary response from users: purchase or non-purchase. This setup makes it harder to accurately measure user preferences. As discussed in the Introduction, mobile games often have limited space for displaying sellable products, which is the case for all three proprietary datasets. This limitation leads to exposure bias in the data. Additionally, placement of different products within the game shop creates positional bias, with some items displayed in more appealing placements while others are not visible on the first screen (Fig. \ref{fig:soda}). Another bias, selection bias, arises from imbalanced product impressions, where certain items—such as conversion offers—are shown to users more frequently, resulting in significantly higher exposure for those items.
\begin{comment}
COAT dataset features 311k biased and 54k unbiased uniform data samples, while YahooR3! contains 12.5k biased and 75k unbiased uniform user-item interactions.
For King's data, Set A features 47.6k biased interactions (for this one, unbiased uniform data is unavailable, which limited the options for debiasing techniques), Set B contains 100k biased and 218k unbiased uniform samples, and Set C includes 980k biased and 1.23 million unbiased uniform interactions. Each dataset exhibited various biases including selection, position, and exposure. 
\end{comment} 

\begin{comment}
For clarity, we give the notation of biased and unbiased uniform datasets.
\begin{itemize}
    \item $D_T$: a biased training set with entries $\{(u_k, i_k, r_k)\}_{1 \leq k \leq |D_T|}$, collected from user interaction history. Here, $u_k \in U$ represents a user, $i_k \in I$ represents an item, and $r_k \in R$ represents a rating, where $U$, $I$, and $R$ are the sets of all users, items, and possible ratings, respectively.
    \item $D_U$: an unbiased uniform set with entries $\{(u_l, i_l, r_l)\}_{1 \leq l \leq |D_T|}$, collected with random assignments.
\end{itemize}
\end{comment} 

\subsection{Selection of Debiasing techniques}
The primary reasoning for the selection of debiasing techniques for this study was based in a literature review, and included the applicability of each method to the specific biases present in the propreitery datasets—namely, selection bias, exposure bias, and position bias. Further, it was imperative to evaluate techniques across two dimensions: those that require randomized datasets and those that do not, as well as to examine methodologies that are agnostic to any particular type of bias. Given the identified biases in the datasets, we adopted several debiasing techniques: (1) \textbf{Matrix Factorisation} (MF) as a baseline model, \textbf{Inverse Propensity Scoring} (IPS), a method that does not require randomized data collection and primarily addresses selection and exposure biases. (2) \textbf{Doubly Robust learning}, that tackles the same biases but, unlike IPS, requires a randomized dataset. And (3) \textbf{AutoDebias} (DR), a bias-agnostic technique that also needs randomized data. Each method was tested across all datasets to evaluate model performance and complexity. We initially applied MF to biased dataset $D_T$ to establish metrics for comparison, we denote our baseline model as \textbf{MF(biased)}, then compared these outcomes with the results from the debiasing methods.

\subsection{Evaluation metrics}
For models' evaluation, we use metrics that assess both predictive power of the models (RMSE and AUC), as well as quality of ranking (NDCG@5) and inequality and diversity in the recommendations (Gini index and Entropy):
\begin{itemize}
    \item \textbf{NDCG@5} assesses the model's ability to rank relevant items in the recommendation list:
\begin{equation*}
\text{NDCG@k} = \frac{\text{DCG@k}}{\text{IDCG@k}}
, \quad
\text{DCG@k} = \sum_{i=1}^{k} \frac{2^{rel_i} - 1}{\log_2(i+1)},
\end{equation*}
 where IDCG@k is the ideal DCG@k and $rel_i$ represents items ordered by their relevance up to position k.

    \item \textbf{RMSE} measures the magnitude of prediction errors of exact rating predictions:
\begin{equation*}
\text{RMSE} = \sqrt{\frac{1}{|R|} \sum_{(u,i) \in R} (\hat{r}_{ui} - r_{ui})^2},
\end{equation*}
where $|R|$ denotes the total number of ratings in the dataset, $\hat{r}_{ui}$ and ${r}_{ui}$ are predicted and true ratings for all user-item pairs $(u,i)$.

    \item \textbf{AUC} reflects how well the model distinguishes between positive and negative interactions:
    \begin{equation*}
\text{AUC} = \frac{\sum_{(u,i) \in D^+_{\text{te}}} \text{rank}_{u,i} - \frac{(|D^+_{\text{te}}| + 1) \cdot |D^+_{\text{te}}|}{2}}{|D^+_{\text{te}}| \cdot (|D_{\text{te}}| - |D^+_{\text{te}}|)},
\end{equation*}
    where $D^+_{\text{te}}$ is the number of positive samples in test set $D_{\text{te}}$, and $rank_{u,i}$ denotes the position of a positive feedback $(u,i)$.
    In experimentation, AUC mainly served as a metric to prevent overfitting and help fine-tunning in validation phase.

    \item \textbf{Gini index} measures inequality in the recommendations distribution. The higher coefficient indicates higher inequality

\[G = \frac{\sum_{i=1}^{n} \left(2i - n - 1\right) \phi_{(i)}}{n \cdot \sum_{i=1}^{n} \phi_{(i)}}\]

Where \( \phi_i \) is the popularity score of the \( i \)-th item, with the scores \( \phi_i \) arranged in ascending order (\( \phi_i \leq \phi_{i+1} \)), and \( n \) represents the total number of items.

    \item \textbf{Entropy} measures the diversity in the distribution of recommended items with higher values indicating higher diversity.
\[
Entropy = -\sum_{i=1}^{n} p_i \log(p_i),
\]

where $n$ is a total number of items u in a dataset and $p_i$ is a probability of an item being recommended.
    
    \end{itemize}
    
    Additionally, we include \textbf{Training Time}, defined as the time required for each model to reach saturation, measured in seconds. This metric provides insights into the computational complexity and the resources required by different methodologies.
\section{Experimentation}

We regard biased data as training set, $D_T$. When it comes to randomized data, following the strategies as mentioned in \cite{liu2020general}, we split it into 3 parts: 5\% for randomised set $D_U$ to help training as required by DR and Autodebias, 5\% for validation set $D_V$ to tune hyper-parameters and incur early-stopping mechanism to prevent overfitting, the rest 90\% for test set $D_{Te}$ to evaluate the model. For conformity reasons, the data split strategy mentioned above is applied to both open datasets and proprietary datasets. For this project, we deploy a training pipeline on Vertex AI \cite{google2023vertexai}, integrating components such as data transformation powered by BigQuery, model training and evaluation, as well as experiment tracking. The training pipeline retrieves data from the data warehouse to train models and produces artifacts that are later integrated into an experiment tracker. By adopting this artifact-based approach, we address the inherent challenge of reproducibility in operationalizing ML projects, as it provides all the necessary components to reproduce experiments. Each experiment is run up to 10 times on Vertex AI with the same hyper parameters, but varying random seeds to get estimation on the variability of the results. 
\begin{comment}
A CI/CD pipeline plays a pivotal role in enhancing software development processes within the industry by facilitating the creation of higher quality code and expediting code deployment. For this project, a training pipeline was implemented on Vertex AI, encompassing components such as data transformation utilizing BigQuery, model training, model evaluation, and experiment tracking. All the experiments were conducted within this framework, ensuring consistency, efficiency, and precision throughout the development lifecycle.
\end{comment}

\section{Experimentation results}

The absolute results of all experiments, including confidence intervals, are presented in Table \ref{tab:performance_metrics}. In this section, we report the percentage improvement of various debiasing techniques compared to the baseline model, which was trained on biased data (MF(biased) model).

\subsection{Open Datasets}
For the \textbf{COAT} dataset, the results show varying degrees of improvement across different metrics (Table \ref{tab:performance_comparison_open_data}). The top performing method (\textbf{AutoDebias}), exhibited the best improvements in RMSE (-5.06\%), AUC (0.39\%) and NGCG@5 (3.73\%) with low changes in Gini (0.16\%) and no improvement in Entropy. \textbf{DR} also provided higher gains in NDCG@5 (2.75\%), and performed better in Gini (-18.88\%) and Entropy (6.16\%), but at a cost of higher RMSE (3.86\%) and lower AUC (-1.57\%). While AutoDebias outperformed other techniques when it comes to improving predictive power of the model (AUC, RMSE), it was not very efficient in terms of Gini and Entropy, and has a significantly higher computational cost. This highlights a trade-off between improved accuracy and increased resource requirements. 

For \textbf{YahooR3!} dataset, again, \textbf{AutoDebias} results in the highest improvement in RMSE (-36.89\%), AUC (1.79\%), NDCG@5 (20.70\%), as well as Gini (-58.15\%) and Entropy (4.26\%), but did so also with dramatically increased computational cost (3216\%). \textbf{IPS} provides a balanced performance with improvements in RMSE (-29.70\%) and Entropy (0.82\%) at a lower computational cost (-22.98\%), making it a practical choice for resource-constrained environments.

\begin{figure}[!htpb]
    \centering
    \includegraphics[width=\linewidth]{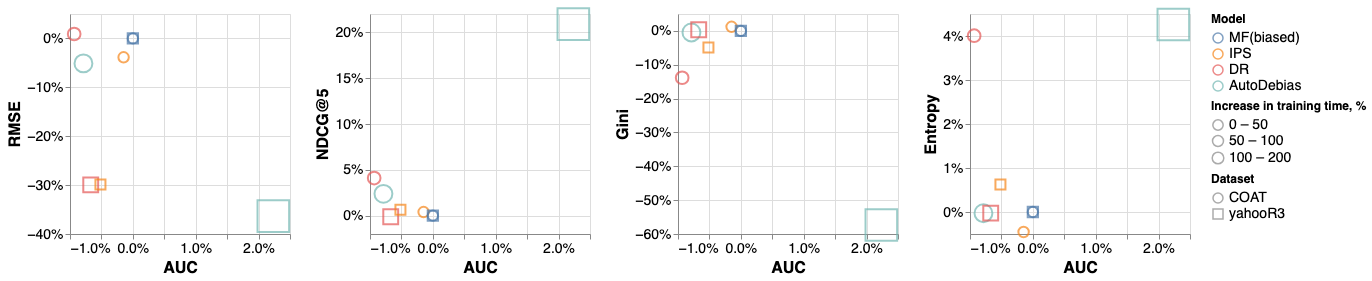}
    \caption{\small{Debiasing results on open datasets (COAT and yahooR3!). The graphs show the percentage change in metrics (AUC, RMSE, NDCG@5, Gini, and Entropy) for various models relative to MF(biased). AUC is plotted against other metrics to demonstrate the trade-off between diversity gains in recommendation systems and potential compromises in predictive power. Different models are represented by colors, training times are indicated by point sizes, and dataset types are distinguished by shapes.}}
    \label{fig:combined_open_data}
\end{figure}
\begin{table}[!htbp]
\caption{Percentage improvement of various models compared to MF(biased) across open datasets. The best results for each metric are highlighted in bold.}
\centering
\begin{tabular}{l l c c c c c c}
\toprule
Dataset & Model & RMSE & AUC & NDCG & Gini & Entropy & Training time (sec) \\
\midrule
\multirow{3}{*}{COAT} 
 & IPS & -2.53\% & -0.26\% & -1.18\% & 0.62\% & -0.29\% & \textbf{8.82\%} \\
 & DR & 3.86\% & -1.57\% & 2.75\% & \textbf{-18.88}\% & \textbf{6.16}\% & 194.12\% \\
 & AutoDebias & \textbf{-5.06}\% & \textbf{0.39}\% & \textbf{3.73}\% & 0.16\% & 0.00\% & 767.65\% \\
\midrule
\multirow{3}{*}{yahooR3!} 
 & IPS & -29.70\% & -0.55\% & 0.73\% & -6.33\% & 0.82\% & \textbf{-22.98}\% \\
 & DR & -30.39\% & -0.83\% & 0.00\% & 1.22\% & -0.12\% & 412.56\% \\
 & AutoDebias & \textbf{-36.89}\% & \textbf{1.79}\% & \textbf{20.70}\% & \textbf{-58.15}\% & \textbf{4.26}\% & 3215.87\% \\
\bottomrule
\end{tabular}
\label{tab:performance_comparison_open_data}
\end{table}

\begin{figure}[!b]
    \centering
    \includegraphics[width=\linewidth]{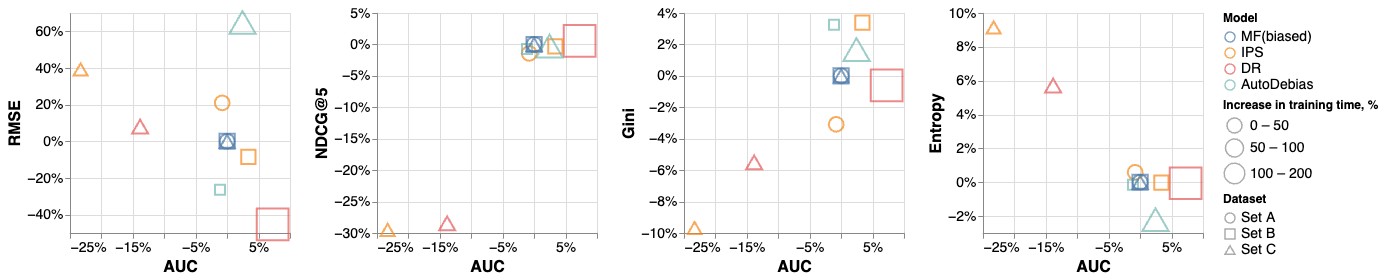}
    \caption{\small{Debiasing results on internal datasets (Set A, Set B and Set C). The graphs show the percentage change in metrics (AUC, RMSE, NDCG@5, Gini, and Entropy) for various models relative to MF(biased). AUC is plotted against other metrics to demonstrate the trade-off between diversity gains in recommendation systems and potential compromises in predictive power. Different models are represented by colors, training times are indicated by point sizes, and dataset types are distinguished by shapes.}}
    \label{fig:combined_king_data}
\end{figure}

\subsection{Internal Datasets}

For the internal datasets, the results are less consistent across the datasets and debiasing techniques (Table \ref{tab:internal_performance_comparison}). This may be due  to the fact that internal datasets employed implicit feedback when collecting data, where user preferences are inferred from their impression and purchase records. This can introduce biases due to the lack of negative samples and overrepresentation of user interactions, potentially skewing the models towards popular items.

\textbf{Set A} is a relatively small dataset (Table \ref{tab:sample_sizes}), and the lack of randomized data limits our options to only using \textbf{IPS}. As a result, some metrics, such as RMSE and AUC, actually worsen (Table \ref{tab:internal_performance_comparison}), which we might accept as a trade-off to achieve better balance in recommendations. However, NDCG@5 also does not improve. On the positive side, IPS enhances diversity metrics, with Gini improving by 3.06\% and Entropy by 0.41\%, while also reducing computational cost by 4.27\%. Overall, applying this method increases model diversity with comparable training time, but comes at the cost of accuracy.

\textbf{Set B} demonstrates substantial improvements with \textbf{DR}, including a 45.40\% reduction in RMSE, a 7.07\% increase in AUC, and gains in NDCG@5 (0.68\%) and Gini (-0.54\%), making the model perform better in both accuracy and diversity. However, this comes at a significant computational cost, increasing training time by 386.46\%. Given the total number of samples being 318k, this leads to a considerably longer training process. \textbf{AutoDebias} ranks second in RMSE improvement (-26.46\%), while \textbf{IPS} shows a positive gain in AUC (3.18\%). However, DR is the only method that consistently improves outcomes of NDCG@5, Gini, and Entropy.

For \textbf{Set C}, the largest dataset with nearly 2.2 million samples, \textbf{AutoDebias} achieves the highest improvement in AUC (2.61\%) and maintains stable NDCG@5. However, it underperforms compared to the baseline and other techniques in RMSE, Gini, Entropy, and training time, which increases significantly by 233.93\%. \textbf{IPS}, on the other hand, delivers poor results in RMSE (39.01\%), AUC (-23.46\%), and NDCG@5 (-29.36\%), but excels in Gini (-9.47\%) and Entropy (9.04\%) without adding to the training time.

\begin{table}[!htbp]
\caption{Percentage improvement of various models compared to MF(biased) across internal datasets. The best results for each metric are highlighted in bold.}
\centering
\begin{tabular}{l l c c c c c c}
\toprule
Dataset & Model & RMSE & AUC & NDCG & Gini & Entropy & Training time (sec) \\
\midrule
\multirow{1}{*}{Set A} 
 & IPS & {20.95}\% & {-0.97}\% & {-1.53}\% & {-3.06}\% & {0.41}\% & {-4.72}\% \\
\midrule
\multirow{3}{*}{Set B} 
 & IPS & -8.61\% & 3.18\% & -0.14\% & 3.29\% & -0.02\% & \textbf{-12.23\%} \\
 & DR & \textbf{-45.40}\% & \textbf{7.07}\% & \textbf{0.68}\% & \textbf{-0.54}\% & \textbf{0.00}\% & 386.46\% \\
 & AutoDebias & -26.46\% & -1.25\% & -0.48\% & 3.26\% & -0.02\% & -63.26\% \\
\midrule
\multirow{3}{*}{Set C} 
 & IPS & 39.01\% & -23.46\% & -29.36\% & \textbf{-9.47\%} & \textbf{9.04\%} & \textbf{-15.50\%} \\
 & DR & \textbf{7.74\%} & -13.76\% & -28.44\% & -5.36\% & 5.47\% & 14.74\% \\
 & AutoDebias & 64.50\% & \textbf{2.61\%} & \textbf{-0.01}\% & 1.72\% & -2.47\% & 233.93\% \\
\bottomrule
\end{tabular}
\label{tab:internal_performance_comparison}
\end{table}

\begin{table}[!htbp]
\caption{Performance metrics across different models and datasets, with 95\% confidence intervals.}
\centering
\begin{tabular}{l l c c c c c c}
\toprule
Dataset & Model & RMSE & AUC & NDCG@5 & Gini & Entropy & Training time (sec) \\
\midrule
\multirow{5}{*}{COAT} 
 & MF (uniform) & 1.00 $\pm$ 0.02 & 0.54 $\pm$ 0.01 & 0.36 $\pm$ 0.02 & 0.64 $\pm$ 0.01 & 4.91 $\pm$ 0.02 & 2.00 $\pm$ 1.60\\
 & MF (biased) & 0.75 $\pm$ 0.01 & 0.77 $\pm$ 0.01 & 0.51 $\pm$ 0.01 & 0.64 $\pm$ 0.04 & 4.9 $\pm$ 0.11 & \textbf{3.40 $\pm$ 1.00}\\
 & IPS & 0.73 $\pm$ 0.01 & 0.76 $\pm$ 0.01 & 0.50 $\pm$ 0.01 & 0.65 $\pm$ 0.04 & 4.89 $\pm$ 0.10 & 3.70 $\pm$ 2.30\\
 & DR & 0.78 $\pm$ 0.02 & 0.75 $\pm$ 0.01 & 0.52 $\pm$ 0.01 & \textbf{0.52 $\pm$ 0.01} & \textbf{5.20 $\pm$ 0.03} & 10.00 $\pm$ 6.90\\
 & AutoDebias &\textbf{ 0.71 $\pm$ 0.01} & \textbf{0.77 $\pm$ 0.02} & \textbf{0.53 $\pm$ 0.01} & 0.64 $\pm$ 0.06 & 4.90 $\pm$ 0.14 & 29.50 $\pm$ 9.6\\
\midrule
\multirow{5}{*}{yahooR3!} 
 & MF (uniform) & 0.73 $\pm$ 0.01 & 0.57 $\pm$ 0.01 & 0.43 $\pm$ 0.01 & 0.41 $\pm$ 0.01 & 6.58 $\pm$ 0.01 & 4.80 $\pm$ 1.20\\
 & MF (biased) & 0.86 $\pm$ 0.01 & 0.73 $\pm$ 0.01 & 0.55 $\pm$ 0.01 & 0.41 $\pm$ 0.01 & 6.58 $\pm$ 0.01 & 60.50 $\pm$ 12.20\\
 & IPS & 0.61 $\pm$ 0.01 & 0.72 $\pm$ 0.01 & 0.55 $\pm$ 0.01 & 0.39 $\pm$ 0.01 & 6.63 $\pm$ 0.02 & \textbf{46.60 $\pm$ 16.10}\\
 & DR & 0.60 $\pm$ 0.04 & 0.72 $\pm$ 0.01 & 0.55 $\pm$ 0.01 & 0.42 $\pm$ 0.01 & 6.57 $\pm$ 0.01 & 310.10 $\pm$ 54.60\\
 & AutoDebias & \textbf{0.54 $\pm$ 0.01} & \textbf{0.74 $\pm$ 0.01} & \textbf{0.66 $\pm$ 0.01} & \textbf{0.17 $\pm$ 0.01} & \textbf{6.86 $\pm$ 0.01} & 2006.10 $\pm$ 1541.00\\
\midrule
\multirow{2}{*}{Set A} 
 & MF (biased) & \textbf{0.82 $\pm$ 0.07} & \textbf{0.54 $\pm$ 0.02} & \textbf{0.56 $\pm$ 0.02} & 0.36 $\pm$ 0.01 & 2.83 $\pm$ 0.01 & 694.30 $\pm$ 163.30\\
 & IPS & 0.99 $\pm$ 0.02 & 0.54 $\pm$ 0.01 & 0.55 $\pm$ 0.01 & \textbf{0.35 $\pm$ 0.02} & \textbf{2.84 $\pm$ 0.02} & 661.5 $\pm$ 85.9\\
\midrule
\multirow{5}{*}{Set B} 
 & MF (uniform) & 0.61 $\pm$ 0.00 & 0.92 $\pm$ 0.01 & 0.97 $\pm$ 0.00 & 0.10 $\pm$ 0.00 & 1.77 $\pm$ 0.00 & 2891.00 $\pm$ 126.90\\
 & MF (biased) & 0.81 $\pm$ 0.06 & 0.89 $\pm$ 0.00 & 0.97 $\pm$ 0.00 & 0.10 $\pm$ 0.00 & 1.80 $\pm$ 0.00 & 2123.90 $\pm$ 441.3\\
 & IPS & 0.74 $\pm$ 0.14 & 0.92 $\pm$ 0.01 & 0.97 $\pm$ 0.00 & 0.10 $\pm$ 0.00 & 1.77 $\pm$ 0.00 & 1864.10 $\pm$ 86.70\\
 & DR & \textbf{0.44 $\pm$ 0.02} & \textbf{0.95 $\pm$ 0.01} & \textbf{0.96 $\pm$ 0.01} & \textbf{0.10 $\pm$ 0.01} & 1.77 $\pm$ 0.00 & 10332.00 $\pm$ 2486.30\\
 & AutoDebias & 0.56 $\pm$ 0.02 & 0.88 $\pm$ 0.01 & 0.96 $\pm$ 0.01 & 0.10 $\pm$ 0.00 & 1.77 $\pm$ 0.00 & \textbf{780.30 $\pm$ 153.70}\\
\midrule
\multirow{5}{*}{Set C} 
 & MF (uniform) & 0.92 $\pm$ 0.04 & 0.25 $\pm$ 0.02 & 0.07 $\pm$ 0.01 & 0.52 $\pm$ 0.01 & 2.52 $\pm$ 0.02 & 775.90 $\pm$ 265.00\\
 & MF (biased) & \textbf{0.62 $\pm$ 0.01} & 0.84 $\pm$ 0.01 & 0.80 $\pm$ 0.01 & 0.65 $\pm$ 0.01 & 2.18 $\pm$ 0.02 & 650.80 $\pm$ 114.70\\
 & IPS & 0.86 $\pm$ 0.06 & 0.64 $\pm$ 0.05 & 0.56 $\pm$ 0.08 & \textbf{0.59 $\pm$ 0.01} & \textbf{2.37 $\pm$ 0.02} & \textbf{549.90 $\pm$ 128.30}\\
 & DR & 0.67 $\pm$ 0.02 & 0.72 $\pm$ 0.05 & 0.57 $\pm$ 0.09 & 0.61 $\pm$ 0.02 & 2.29 $\pm$ 0.05 & 746.70 $\pm$ 140.00\\
 & AutoDebias & 1.02 $\pm$ 0.03 & \textbf{0.86 $\pm$ 0.04} & \textbf{0.78 $\pm$ 0.02} & 0.66 $\pm$ 0.02 & 2.12 $\pm$ 0.04 & 2173.20 $\pm$ 1826.10\\
\bottomrule
\end{tabular}
\label{tab:performance_metrics}
\end{table}

\section{Conclusion and Future work}

%Implementing more accurate, less biased models helps preventing the perpetuation of negative feedback loops and overexposure of certain items caused by segmentation heuristics in retraining data. This also improves data quality for fine-tuning models. A RS that diversifies exposure enhancing user experience by not limiting visibility to popular items only.

Implementing more accurate and less biased models is crucial to avoiding the perpetuation of negative feedback loops and the overexposure of certain items caused by segmentation heuristics in retraining data. This approach also enhances data quality, which is essential for fine-tuning models. A recommender system that diversifies content exposure improves user experience by ensuring that visibility is not limited to only the most popular items. In our experiments, Inverse Propensity Scoring (IPS) stands out for its simplicity and model-agnostic nature, requiring no randomized data collection and fewer training epochs. However, the improvements it offers are somewhat limited. AutoDebias excels in improving accuracy metrics, but at substantially higher computational costs and sometimes poorer performance in Gini and Entropy. DR still offers strong improvement in observed metrics, including Gini and Entropy. So while each debiasing method has its own trade-offs, significant performance gains still depend on the challenging task of collecting randomized datasets, as highlighted in our introduction. Potential future work includes:  (1) adopting online reinforcement learning approach such as Multi-Armed Bandit (MAB) \cite{felicio2017multi,wang2017biucb,wang2018online} for data collection, including contextual bandit models, (2) developing and testing combined debiasing models which can combine strengths of different debiasing techniques to mitigate various biases simultaneously while optimizing for computational efficiency.

%\\subsection{Summary}
%\\textbf{AutoDebias} excels in improving accuracy metrics, but at substantially higher computational costs and sometimes poorer performance in Gini and Entropy. \textbf{DR} still offers strong improvement in observed metrics, including Gini and entropy, and is also rather expensive computationally, but much less so. \textbf{IPS} provides a balanced performance with lower computational cost, making it a potentially more practical choice for scenarios, where such factors like computational cost and balanced performance are crucial.

%\begin{credits}
%\subsubsection{\ackname} The authors are grateful for the support provided by King while working on the study and the manuscript. In addition, we would like to give special thanks to the teams at King for their help and support: ML Special Projects, AI Labs, and CCS IAP\&E.

%\end{credits}

\end{document}